\documentclass[letterpaper, 10 pt, conference]{ieeeconf}  %

\IEEEoverridecommandlockouts                              %

\usepackage[backend=bibtex,bibstyle=ieee,citestyle=numeric-comp,maxbibnames=3,maxcitenames=3,doi=false,url=false,eprint=false]{biblatex}
\usepackage{graphics} %
\usepackage{graphicx}

\usepackage[svgnames]{xcolor}
\usepackage{amsmath}
\usepackage{svg}
\usepackage{subcaption}
\usepackage{amssymb}

\title{\LARGE \bf
Leveraging Swarm Intelligence to Drive Autonomously: \\ A Particle Swarm Optimization based Approach to Motion Planning
}

\author{Sven Ochs$^{1*}$, Jens Doll$^{1*}$, Marc Heinrich $^{1*}$, Philip Sch\"orner$^{1}$,\\
Sebastian Klemm, Marc Ren\'{e} Zofka$^{1}$ and J. Marius Z\"ollner$^{1,2}$%
\thanks{$^{1}$ Department of Technical Cognitive Systems, FZI Research Center for Information Technology, Germany.
	{\tt\small \{surname\}@fzi.de}}%
\thanks{$^{2}$ Karlsruhe Institute of Technology (KIT), Germany.}
}

\begin{document}

\maketitle
\thispagestyle{empty}
\pagestyle{empty}

\begin{abstract}

Motion planning is an essential part of autonomous mobile platforms. A good pipeline should be modular enough to handle different vehicles, environments, and perception modules. The planning process has to cope with all the different modalities and has to have a modular and flexible design. But most importantly, it has to be safe and robust. In this paper, we want to present our motion planning pipeline with particle swarm optimization (PSO) at its core. This solution is independent of the vehicle type and has a clear and simple-to-implement interface for perception modules. Moreover, the approach stands out for being easily adaptable to new scenarios. Parallel calculation allows for fast planning cycles. Following the principles of PSO, the trajectory planer first generates a swarm of initial trajectories that are optimized afterward. We present the underlying control space and inner workings. Finally, the application to real-world automated driving is shown in the evaluation with a deeper look at the modeling of the cost function. The approach is used in our automated shuttles that have already driven more than 3.500 km safely and entirely autonomously in sub-urban everyday traffic.

\end{abstract}

\section{INTRODUCTION}

Motion planning is a fundamental problem in robotics and autonomous systems. It requires generating feasible and optimal trajectories for a robot or vehicle to navigate its environment while avoiding obstacles and adhering to predefined constraints. In the context of automated vehicles, this involves the generation of optimal trajectories and control commands to navigate a dynamic and often uncertain environment while ensuring safety, efficiency, and compliance with traffic regulations.

It is a multifaceted problem encompassing various considerations, such as real-time sensor data processing, obstacle detection and avoidance, lane keeping, and route optimization. The successful execution of these tasks hinges on the ability to make rapid and intelligent decisions in a highly dynamic and often unpredictable environment. This environment becomes unpredictable through interactions with other road users like cars, bicyclists, and pedestrians, as well as various road, weather, and lighting conditions. This poses challenges, especially for the motion planning pipeline regarding re-planning, compute efficiency and comfortable ride. 

Particle Swarm Optimization (PSO), inspired by the collective behavior of social organisms, offers an elegant solution to the intricate motion planning problem in automated vehicles. In PSO, a population of particles represents potential solutions in a multi-dimensional search space. These particles adaptively explore and exploit the solution space by iteratively adjusting their positions based on their own experiences and peers' experiences. By leveraging this cooperative optimization process, PSO seeks to identify the optimal trajectory that minimizes a defined cost function.
In this work, we chose to use particle swarm optimization to find the optimal solution. In the context of automated driving, real-time constraints are imposed on every function to ensure an upper bound on the reaction time. Therefore, an anytime-capable process like PSO is required. Furthermore, soft as well as hard constraints need to be considered. PSO is also able to find the globally optimal solution for any problem when initialized correctly. Last but not least, the function to optimize is nonlinear. Particle swarm optimization was selected because it excels in all these criteria while knowing about the downsides of not having guarantees to find the optimal solution with a limited number of particles and iterations.

\begin{figure}[t]
	\centering
		\includegraphics[width=\columnwidth]{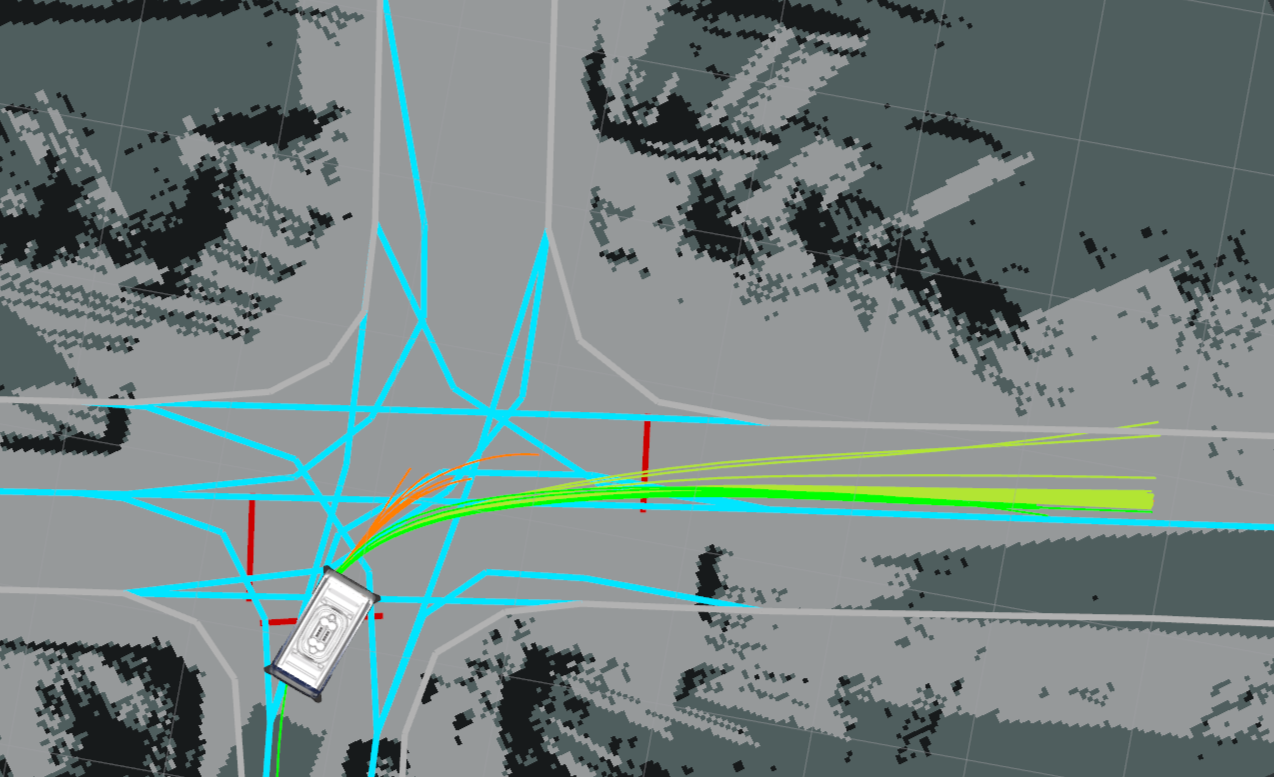}
		\caption{Our PSO planer implementation in action in a sub-urban area. The orange line indicates the particles used for the initialization and the green represent the trajectories after optimization.}
		\label{fig:pso_intro}
\end{figure}

In Sec.~\ref{sec:architecture}, we present our trajectory optimization approach. In Sec.~\ref{sec:realization} the details regarding our realization from the PSO and its inputs are described, followed by the evaluation in Sec.~\ref{sec:evaluation}. We conclude our work in Sec.~\ref{sec:conclusion}.

\newpage
\section{RELATED WORK}
Planning safe and feasible trajectories for automated vehicles in real traffic requires a planning technique that is able to consider both the vehicle's restricted maneuverability and dynamic constraints, as well as constraints inflicted through the environment like keeping to roads or avoiding obstacles.
Moreover, complex scenarios require the handling of multi-modal problems.

In \cite{ziegler2014}, an optimization based approach for on-road driving of an autonomous vehicle is proposed. However, the sequential quadratic programming approach has difficulties dealing with hard constraints like collisions due to the need to calculate gradients. Furthermore, it requires effortful representations to formulate the driving task as a convex problem. Therefore, it is unsuitable to handle multi-modal scenarios.

Search-based planners like RRT \cite{lavalle1998rapidly}, optimal versions like the RRT* or kinodynamic RRT* \cite{webb2012kinodynamic,klemm2015rrt}, on the other hand, bring the advantage of completeness and optimality.
For a review of sampling-based planners, the reader is referred to \cite{Elbanhawi2014}, and for a review on motion planning algorithms used in the context of automated driving to \cite{Gonzales2016}.

Nietzschmann et al.\cite{nietzschmann2018trajectory}  solve the trajectory planning problem using the
formulation of an optimal control problem (OCP)\cite{ocp}. This algorithm is targeted specifically for the kinematic single-track model. The paths, lanes, and boundaries have to be provided in a fixed set of (x,y) points. This is a drawback especially in changing environments and moving road boundaries due to for example construction work or parked vehicles. 

Learning base approaches like ChauffeurNet \cite{bansal2018chauffeurnet} demonstrate how trajectories can be determined using machine learning.
While these approaches promise a high degree of generalization, they lack traceability and safety constraints.

A hybrid approach to use the strengths of sampling-based planners and a fast optimization can be achieved by utilizing particle swarm optimization \cite{kennedy2010particle}.
The method is also capable of handling hard constraints \cite{hu2002solving}.
In the context of mobile robots, this can be barriers or collisions with obstacles.
An overview on variants of the particle swarm optimization is given in \cite{poli2007} and a short description is given in the next section.

\section{TRAJECTORY OPTIMIZATION}
\label{sec:architecture}

Every motion pipeline's main problem is finding the optimal trajectory regarding its constraints and cost terms. The trajectory depends on the current arbitrary environment comprised of the inputs from localization, perception, and maneuver decisions. The problem is formulated as an optimization problem in which either the minimum costs or the maximum reward is to be determined. With input data $x$ within the domain $D$, the optimization of function $f$ can be formulated as 

\begin{equation}
    \arg \min_x f(x)
\quad \textrm{with} \quad
x \in D.
\label{eq:allgemeine_optimization}
\end{equation}

We utilize the PSO algorithm for our trajectory calculations. Its unique feature lies in its capacity to function without imposing substantial assumptions about the nature of the optimization problem it confronts, enabling it to explore extensive arrays of potential solutions. PSO's remarkable attribute is its detachment from the need for the gradient of the problem under consideration, unlike traditional optimization techniques like gradient descent and quasi-Newton methods. This independence from differentiability opens doors for PSO to address non-differentiable problems. Moreover, large parts of the PSO calculations are highly parallelizable. These advantages outweigh the drawback that PSO cannot guarantee to find the global optimum. In the context of motion planning, an approximation of the optimal solution is often sufficient.

The properties of PSO can be formulated as follows:
\begin{itemize}
    \item scale-invariant: $f(x) \approx c \cdot f(x) \in c > 0$
    \item offset invariant: $f(x) \approx c + f(x)$
    \item gradient scale invariant: $\nabla f(x) \approx c \cdot \nabla f(x) \in c > 0$
    \item Insensitive to the magnitude of the gradient
    \item detachment from the need for the gradient
    \item $f(x)$ must be defined on the on the input data
\end{itemize}

These features are the main motivation why we use PSO for motion planning. This enables us to define different cost terms without the need to engineer complicated representations to generate gradients or convexity. Thus, we gain the freedom to design cost terms to handle various scenarios in suburban and urban areas. Moreover, the beneficial computational properties allow for short turnaround times,  which are necessary for high interaction rates with dynamic obstacles and changing environments.

The basic procedure of PSO is divided into the following three steps.

\subsubsection{Initialization}
First, the particle swarm has to be initialized. Each particle $\vec{x}$ represents a possible solution to the optimization problem, whose fitness can be derived from the given cost function $f$.
In addition to its initial state, each particle is assigned an initial velocity, which describes its movement through the search space.
All of the particles know their best state in the search space found so far $\vec{x}_\mathrm{p}$ as well as the corresponding fitness value.
Also fitness and position of the swarm's best particle $\vec{x}_\mathrm{g}$ are known.
To ensure finding the optimal solution, the search space needs to be covered thoroughly. If prior knowledge is available, the distribution can be guided using heuristics. Otherwise, the particles can be distributed randomly within the search space.

\subsubsection{Movement of the swarm}
After the initial population of the particle swarm has been generated, the particles move through the search space.
Their movement depends on their current state $\vec{x}$ and the best position $\vec{x_\mathrm{g}}$ within the entire swarm found so far.
The state of the particles is recalculated after each optimization step as follows
\begin{equation}
	\vec{v}_{t+1} = w_\mathrm{v} \cdot \vec{v}_t + w_\mathrm{p} \cdot u_1 \cdot (\vec{x}_{\mathrm{p},t} - \vec{x}_t) + w_\mathrm{g} \cdot u_2 \cdot (\vec{x}_{\mathrm{g},t} - \vec{x}_t) \, ,
	\label{eq:updateParticleVelocity}
\end{equation}

with $w_\mathrm{v}$, $w_\mathrm{p}$ and $w_\mathrm{g}$ being constant weight factors and
$u_i \sim U(0,1)$ uniformly distributed random numbers.\\
The velocity is used to update the particle state
\begin{equation}
	\vec{x}_{t+1} = \vec{x}_t + \vec{v}_{t+1} \, .
	\label{eq:updateParticlePosition}
\end{equation}

After the particles have been updated, they are checked for compliance with the given constraints. Valid particles will have their fitness value reassessed according to their new state within the search space.
Invalid particles are kept in the swarm, but their fitness value is not updated.
If a new individual or swarm minimum has been reached, $\vec{x}_\mathrm{p}$ and $\vec{x}_\mathrm{g}$ are updated as well.

\subsubsection{Termination}
The optimization process is stopped if one of the termination criteria is met.
These include reaching a maximum number of iterations, a specific fitness value, or exceeding a given compute time limit.
Another reason to terminate the optimization process can be the loss of diversity within the swarm. In that case, the particles are barely moving, and there is little or no improvement in their fitness values.

\subsection{Representation of trajectories}

A particle of the swarm represents a single trajectory $T$.
The trajectory $T$ is built from a sequence of SE2 poses with equidistant time intervals. Starting at $t=0$, there is one SE2 pose for each $t_{i+1}=t_i + \Delta t$.
A SE2 sate $\Vec{p}= \{x, y, \theta \}$ contains the position $(x,y)$ and orientation or yaw angle $\theta$.
A trajectory results to $T= \{\Vec{p}_0,\dots,\Vec{p}_{n-1} \}$ for a trajectory of size $n$.

We model our motion planning problem using a trajectory $T$ comprised of SE2 poses with a discrete timestamp $t_i$:

\begin{equation}
T :
\begin{cases}
\mathbb{N} \rightarrow SE2 \\
t_i \rightarrow \Vec{p}(i)
\end{cases}
    \label{eq:pso_basic_problem}
\end{equation}

This design decision is made to fuse other modules like perception or localization into the planning process. Since PSO imposes so few preconditions, this process can easily be expanded to a time-independent trajectory design. Currently, we focus on a time discrete approach. This also simplifies the transitions from one time step $t$ to the following one $t+1$. This transition is described by so-called controls $c$. The transition from state $\Vec{p}_{t}$ to $\Vec{p}_{t+1}$ can thus be described by

\begin{equation}
\Vec{p}_{t+1} = f_c(\Vec{p}_{t},c_t).
    \label{eq:transition_function}
\end{equation}

The transition function $f_c$ takes the control value and the current state as input. This leads to the new state $\Vec{p}_{t+1}$. Through that, a complete trajectory can be expressed by a starting state $\vec{p}_0$, an ordered sequence of control values $c_0,\dots,c_{n-1} \in C$ and a transition function $f_c: S \times C \mapsto S$ within the state space $S$. The general form of control in SE2 can be described as offsets $\Delta$ in $x$, $y$, and $\theta$ as seen in (\ref{eq:general_control}).

\begin{equation}
c(\cdot) = {\Delta x, \Delta y, \Delta \theta}
    \label{eq:general_control}
\end{equation}

\begin{figure}[t]
	\centering
		\includegraphics[width=\columnwidth]{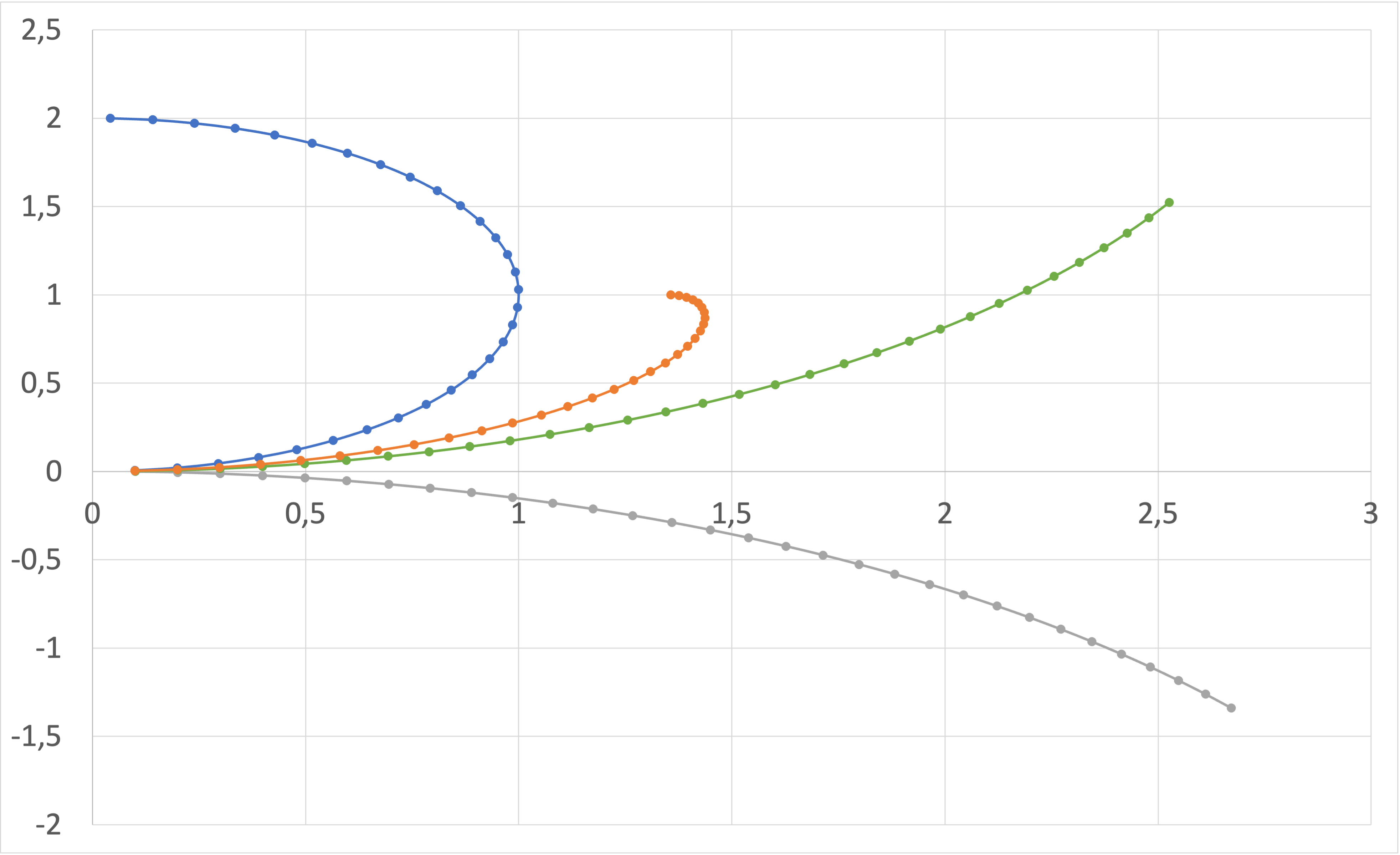}
		\caption{Interpolation between two trajectories (blue and grey). The green one is interpolated using curvature controls, the orange one with Cartesian interpolation.}
		\label{fig:interpolate_in_control_space}
\end{figure}

This representation as a transition function also enables the kinematically correct interpolation between different particles, as needed in \ref{eq:updateParticleVelocity}. If the interpolation would be done in the Cartesian space, there would be unfavorable results, as visualized in Fig.~\ref{fig:interpolate_in_control_space}.

In this figure, we present two different interpolation approaches. In blue and grey, two particles are depicted, representing a constant curvature trajectory with $\kappa=1$ and $\kappa=-0.3$. The orange trajectory is interpolated using the Cartesian coordinates, and the green one uses the interpolation in control space. One can observe that the orange path deviates from constant curvature, which is typically anticipated when interpolating between two constant curvature trajectories. The interpolation in the control space yields the expected result with $\kappa=0.35$. For this reason, the representation of trajectories in the control space is used. Also, we use the polar-coordinate-control (\ref{eq:polar_control}), since it allows smoother interpolation than the Cartesian interpolation.

\begin{equation}
c(l, \kappa) = \Biggl ( l \cdot \cos(\frac{1}{2} \kappa \cdot l), l \cdot \sin(\frac{1}{2} \kappa \cdot l), \kappa \cdot l \Biggr )
    \label{eq:polar_control}
\end{equation}

The factor of $1/2$ can be geometrically justified by the fact that a rotation of 180° represents a semicircle, but when viewed from the origin, it only describes an angle of 90°. 

\subsection{Constraints}

To ensure a trajectory is safe and feasible, constraints must be met during the optimization. For that, we use inequality constraings $g(x)$:

\begin{equation}
    \underset{x}{\arg \min} f(x) \quad \textrm{with} \quad x \in D ~ \textrm{,} ~ g(x) \ge 0.
\end{equation}

Common boundary conditions can be divided into internal and external constraints. External constraints describe boundary conditions regarding the environment. First of all, trajectories have to be free of collisions. While cost terms can motivate such behavior, they fail to enforce it. Therefore, a hard constraint for a minimum distance to these obstacles ensures that no collision will occur. The distance to obstacles and collisions are checked using a similar method as in \cite{ziegler2014}. To improve computational efficiency, circles are used to approximate the ego-vehicle's collision model. Surrounding obstacles are modeled using polygons. The same model ensures the vehicle does not leave the determined driving area.
Internal constraints are related to the dynamic properties of the vehicle. Those are motivated by comfort (accelerations, jolt, and yaw rates) and physical aspects (kinematic and dynamic vehicle limits). The latter include drivetrain limitations and actuator limits (steering speed and acceleration).
Furthermore, the vehicle needs to stick to the current speed limit.

\subsection{Costs}

In this section, we take a closer look at the fitness of the particles. The fitness is calculated as the weighted sum over various cost terms, each representing a different desired behavior. The cost functions used in the optimization process are as follows:
\begin{itemize}
    \item velocity: deviations from the desired velocity
    \item acceleration: limits the dynamics to the defined maximum acceleration
    \item jolt: cost term to improve comfort by minimizing acceleration variations
    \item driving-area: make the vehicle keep a distance to the boundaries of the driving area
    \item orientation: encourages the planner to drive in the same direction as the lane leads
    \item yaw-rate: hinders the planner from changing direction fast and motivates it to drive smoothly, thus improving comfort
    \item halting: enables the planner for smooth braking for stop lines
    \item obstacle clearance: increases the costs when the ego-vehicle passes close to obstacles
    \item lateral bias: keep to the lateral bias set by the maneuver planner for, e.g. stopping at the outer side of the road
\end{itemize}

The cost terms above are all incorporated in the planner simultaneously. Some of them are orthogonal to each other. Therefore, an equilibrium must be found. Since some cost terms, like obstacle clearance, are more important than comfort, this equilibrium must be defined with care.

\subsection{Adaptations}
The cost terms are an important modeling task since they influence the optimization process. However, advanced search space sampling enables the optimizer to find better solutions. In the standard implementation of the constrained PSO, all particles are no longer inspected if they violate at least one constraint. Thus, it does not optimize well along complex constraints. To improve this, our policy deviates from this behavior by further propagating the particles even if they violate multiple constraints.
The solution can further be improved by applying advanced sampling algorithms. This enables the optimizer to start from a set of better initial particles.

Firstly, we loop back the past best solution $x_{g (i-1)}$ as one of the new initial particles. Other previous particles, that are still valid within the updated environment, are also taken over. However, 30\% of those particles are randomly dropped to aid exploration and responsiveness to situational changes. All remaining particles are sampled according to heuristics to guide the search towards valid parts of the search space. It is mandatory to yield a certain minimum of valid particles for the initial swarm. To improve computation time, we seek to increase the rate of valid samples by limiting the sampling space with the constraints. During the sampling process, it can be ensured that only valid successive states can be generated by using the inverse kinematics while complying with all internal constraints. For external constraints, a path search planner such as A* can be used to guide the sampling. However, we found that the computational overhead did not yield a net time gain.
Furthermore, machine learning algorithms can provide learned heuristics for improved initialization, as shown in \cite{schoerner2020optimization}.

Since our planer is deployed in real autonomous vehicles driving in everyday traffic, we constrain our trajectories to ensure continuity. In continuous planning, trajectories are truncated according to the time since the last trajectory. To avoid introducing discontinuities, the trajectory uses a non-changeable horizon unchanged from the previous iteration. This horizon is defined by the turnaround time plus the planning pipeline latency. This avoids creating trajectories infeasible to execute by real-world vehicles, especially with internal constraints.

Because our problem is formulated using a time-discrete SE2 state sequence, certain dynamic and kinematic values of higher order must be calculated using multiple states. We use a prefix ranging into the past to ensure a well-defined initial state at the current vehicle position. Moreover, it ensures positional continuity at the ego-vehicle's current position. The prefix is visualized as dark-green boxes in Fig.~\ref{fig:neckar_rviz} and Fig.~\ref{fig:freiburger_rviz}.
\section{REALIZATION}
\label{sec:realization}

\begin{figure}[t]
	\centering
		\includegraphics[width=\columnwidth]{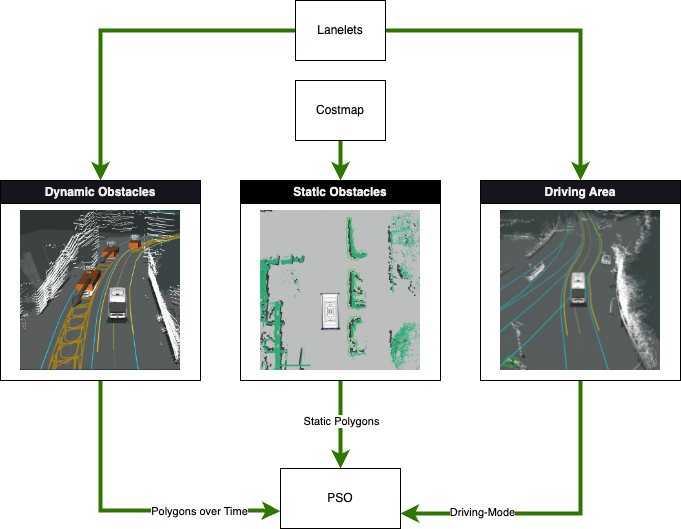}
		\caption{The inputs of the PSO. The environment is modeled by static and dynamic obstacles represented as polygon features. Maneuver decisions are passed via the driving area and the driving mode. }
		\label{fig:overview}
\end{figure}

In this section, we would like to present the surrounding software environment and the inputs the PSO receives in our implementation. The motion pipeline is implemented in a top-down approach where the abstraction level decreases between each layer. The top level represents the mission. A mission consists of a start and goal position encoded in GPS positions. After knowing the global goal of our mission, we calculate the route utilizing the lanelet format \cite{bender2014lanelets}. The maneuver layer extracts a lane envelope called the driving area, where the planner id allowed to plane freely. Using a frenet system, further maneuver decisions are derived. These constraints are encoded in a semantic model called driving mode, which contains, among others, a desired velocity as well as dynamic non-passable constraints or stop lines. For the environment model of the planner, we separate the dynamic from the static obstacles. The static obstacles are encoded in a occupancy grid, the so-called costmap \cite{ros}. From the costmap, we extract static obstacles as polygons as planner input. The result can be seen as green outlines around the detected obstacles, displayed as black pixels in the costmap in Fig.~\ref{fig:overview}. Dynamic obstacles are directly processed as polygons by the perception preprocessing and passed as such into the planner.

 The overall visualization of the planning process is presented in Fig.~\ref{fig:neckar_rviz} and Fig.~\ref{fig:freiburger_rviz}. In cyan, the lanelet map is presented, which is the basis for the driving area, presented in yellow, which the maneuver decision can dynamically change. The green and orange line originating from the center of the vehicle represents the swarm of the PSO. The orange particles are the initial ones used for optimization. The final swarm is presented as green lines. The best particle is then used for the output trajectory. In the different green shades, the final calculated trajectory is presented.

\section{EVALUATION}
\label{sec:evaluation}

\begin{figure*}[t]
    \begin{subfigure}[b]{0.25\textwidth}
        \includegraphics[width=\textwidth]{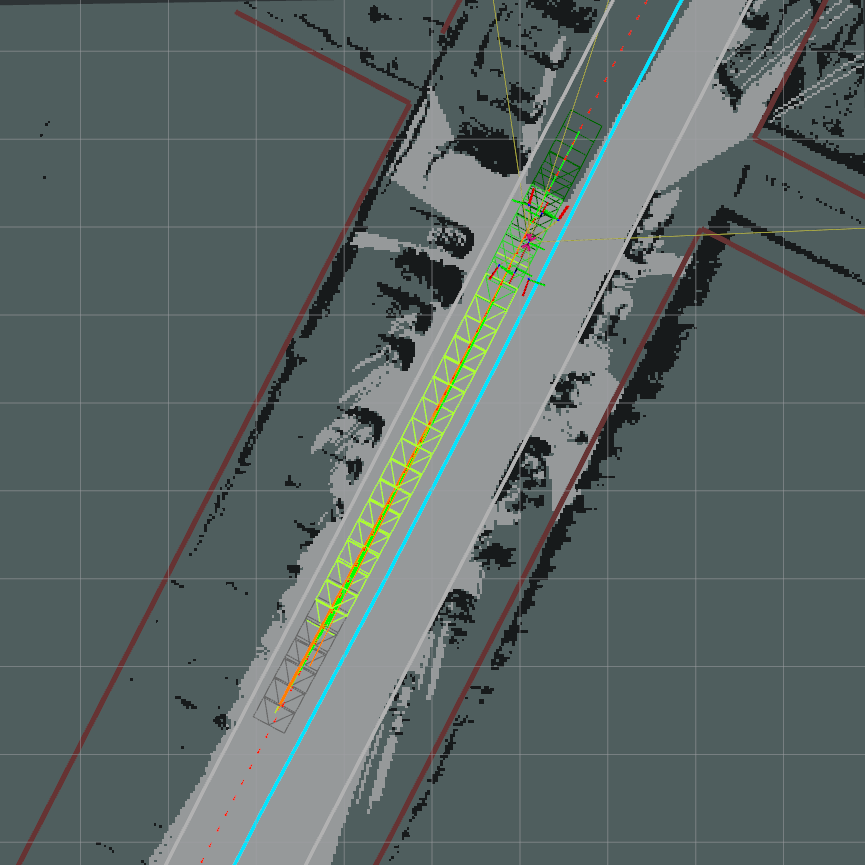}
        \caption{
        Visualization of scenario I}
        \label{fig:neckar_rviz}
    \end{subfigure}%
    \hspace{0.05\textwidth}
    \begin{subfigure}[b]{0.67\textwidth}
        \includegraphics[width=\textwidth]{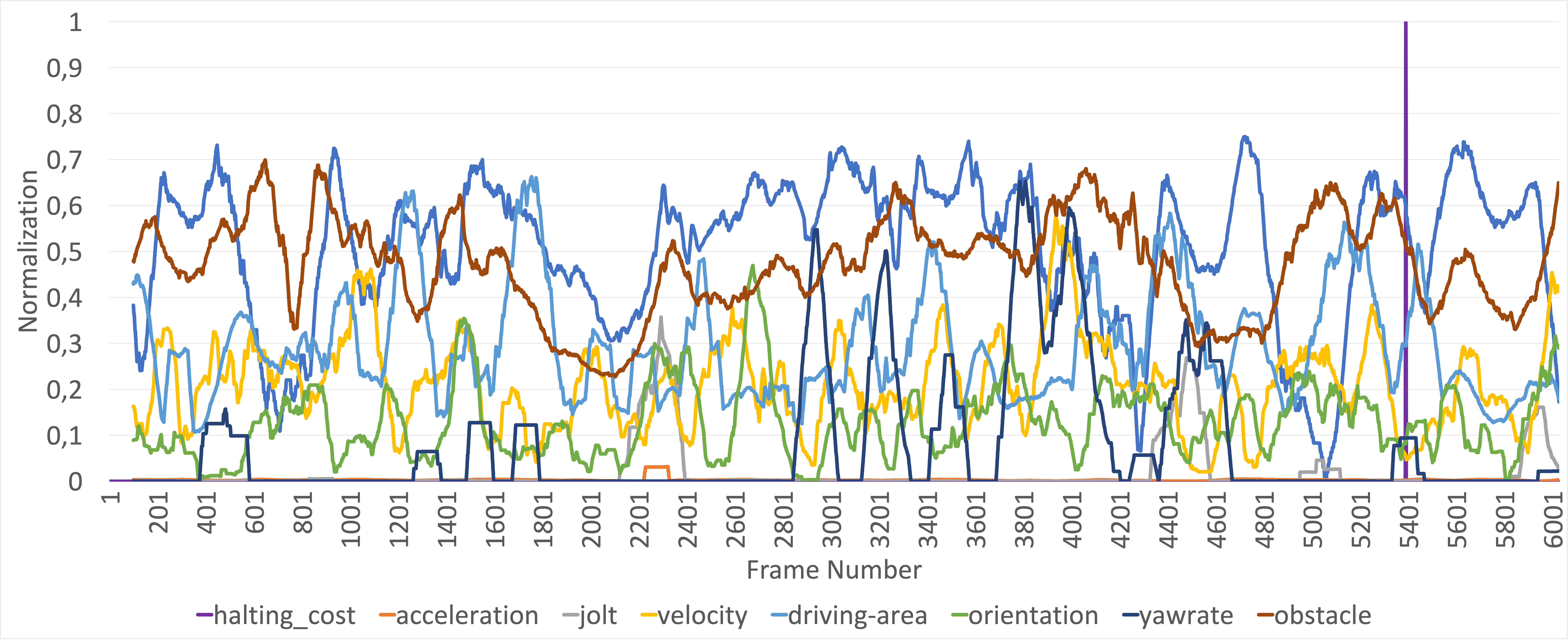}
        \caption{
        Individual cost terms for the best trajectory of scenario I
        }
        \label{fig:neckar_cost}
    \end{subfigure}%

    \vspace{0.5cm}
    
    \begin{subfigure}[b]{0.25\textwidth}
        \includegraphics[width=\textwidth]{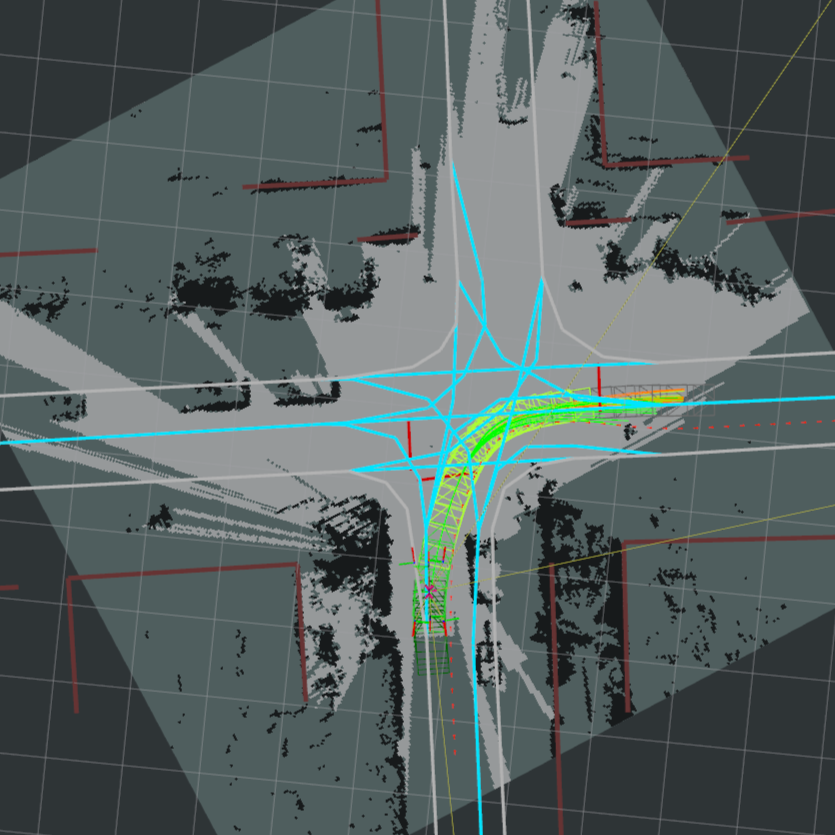}
        \caption{Visualization of scenario II}
        \label{fig:freiburger_rviz}
    \end{subfigure}%
    \hspace{0.05\textwidth}
    \begin{subfigure}[b]{0.67\textwidth}
        \includegraphics[width=\textwidth]{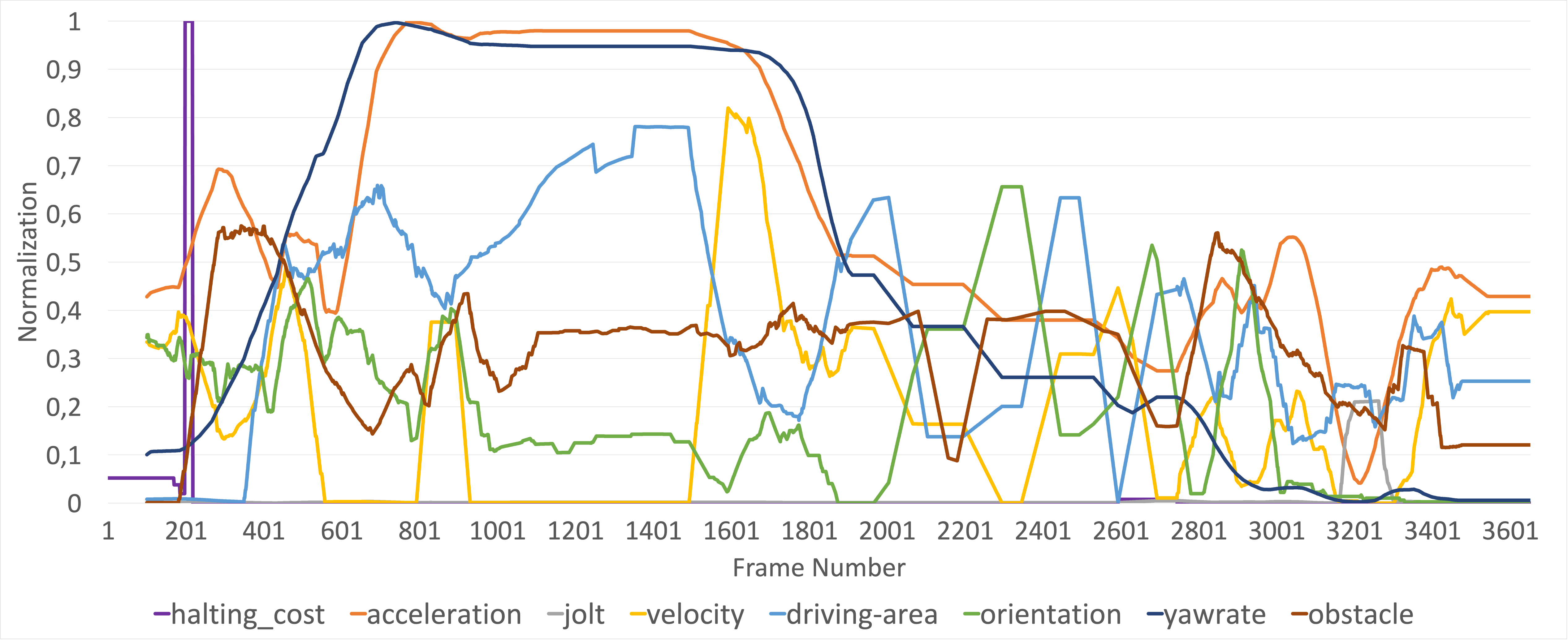}
        \caption{%
        Individual cost terms for the best trajectory of scenario II
        }
        \label{fig:freiburger_cost}
    \end{subfigure}%

    \vspace{0.5cm}
    
    \begin{subfigure}[b]{0.42\textwidth}
		\includegraphics[height=4.8cm]{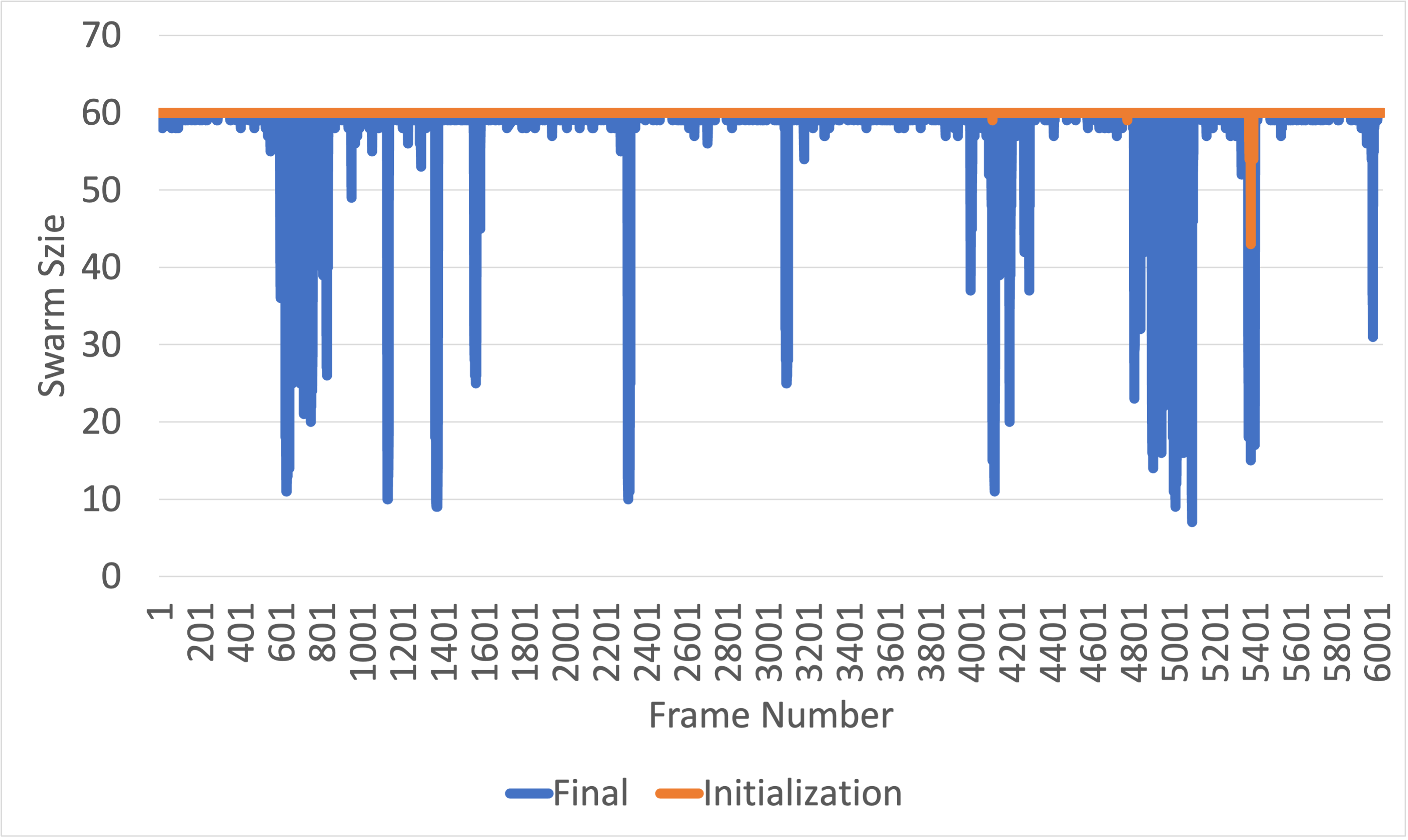}
        \caption{
        Number of valid particles of scenario I
        }
		\label{fig:neckar_timing}
\end{subfigure}%
    \hspace{0.05\textwidth}
\begin{subfigure}[b]{0.42\textwidth}
		\includegraphics[height=4.8cm]{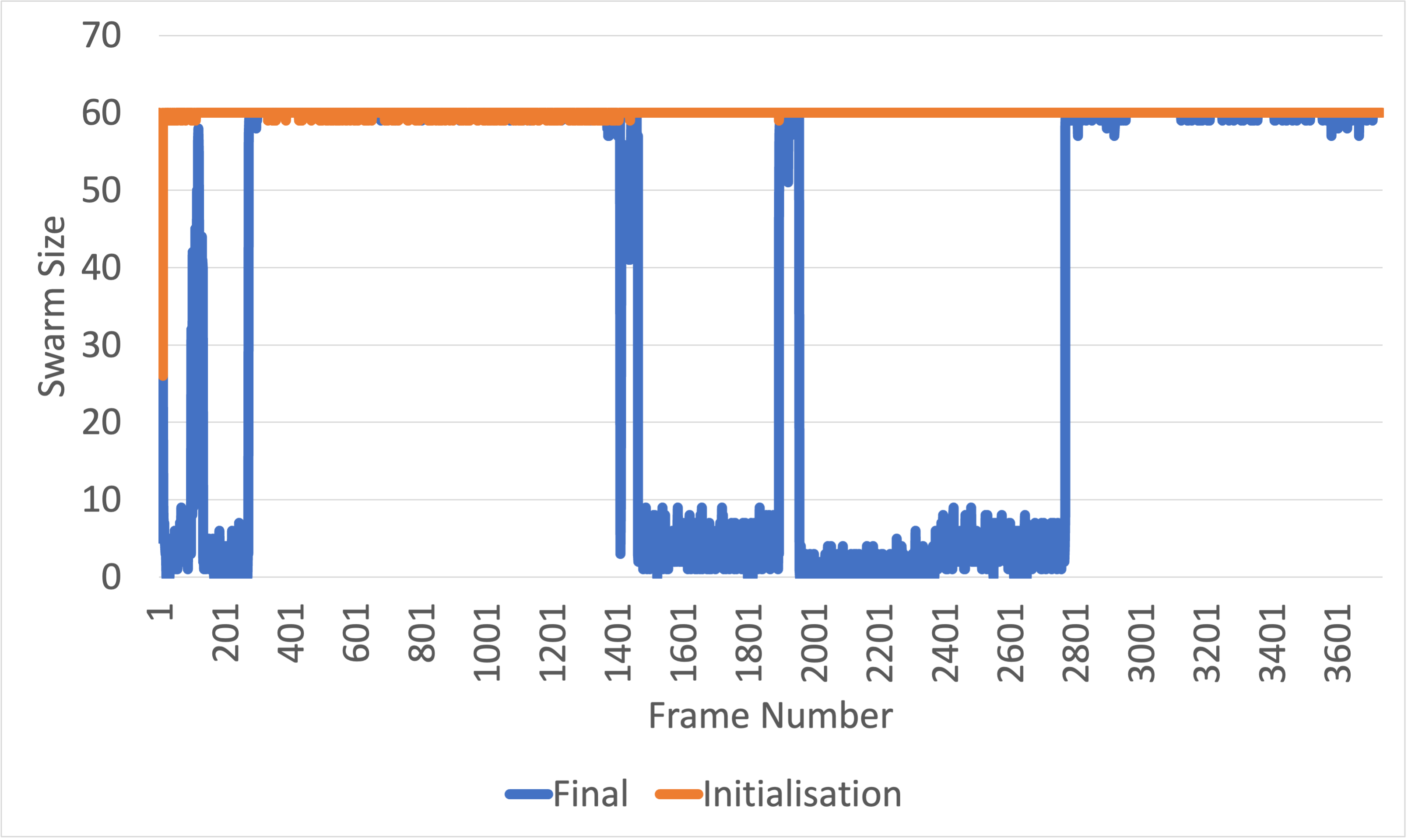}
        \caption{
        Number of valid particles of scenario II
        }
        \label{fig:freiburger_timing}
\end{subfigure}

    \caption{Depiction and statistics of scenarios I (images \subref{fig:neckar_rviz},\subref{fig:neckar_cost},\subref{fig:neckar_timing}) and II (images \subref{fig:freiburger_rviz},\subref{fig:freiburger_cost},\subref{fig:freiburger_timing}).
    The scene with current trajectory as vehicle contours (green), the particles (orange) and underlying occupancy grid is shown in   (\subref{fig:neckar_rviz}) and (\subref{fig:freiburger_rviz}).
    The normalized individual cost terms over time are depicted in  (\subref{fig:neckar_cost}) and (\subref{fig:freiburger_cost}).
    Finally, the valid particles over time are given in (\subref{fig:neckar_timing}) and (\subref{fig:freiburger_timing}).
    The number of valid particles for the initialization phase is presented in blue and after the optimization in orange.
    }
\end{figure*}

In the following, we evaluate our approach in two different real traffic scenarios. All the evaluations are conducted on an AMD Ryzen™ Threadripper™ 3960X with five cores dedicated to calculating the particles. Our swarm comprises 60 particles, each representing a trajectory with 25 points with a time interval of $\Delta t = 0.3\,$s, resulting in a total horizon of 7.2\,s. The turnaround time for a maximum of 50 iterations is at 11\,ms with only a few exceptions at a maximum 23\,ms. This enables us to easily reach planning frequencies above 20\,Hz even on slower hardware. This allows the use of the planner on all autonomous mobile platforms. For evaluation purposes, we sample the cost terms with 10\,Hz.

Fig.~\ref{fig:neckar_rviz} and Fig.~\ref{fig:freiburger_rviz} show two different driving scenarios. The first one in Fig.~\ref{fig:neckar_rviz} is a main street in a suburban area with parking spaces alongside the road. In this street, reaching the maximum allowed speed is desirable while keeping a safe distance from all obstacles nearby. For both permission and perceived safety, the safe distance between the ego vehicle and the obstacles has to increase with increasing velocity. This behavior is reflected in the cost terms of Fig.~\ref{fig:neckar_cost}. The cost terms for the driving area and the obstacle avoidance are constantly highest, followed by the velocity. This distribution is expected and shows that the terms are tuned correctly. Terms regarding cornering, like yaw rate, are negligible as expected because there is no need for steering.

The planner can show its capabilities in the second scenario, as seen in Fig.~\ref{fig:freiburger_rviz}. The ego vehicle is driving from a side street into a junction. Additionally, a vehicle is parked around the corner blocking the ego-vehicle's lane. The planner, therefore, has to cope with multiple challenges. First, he has to react to the right of way through a halt line, reflected by the abrupt appearance of the halting cost term at frame 200. Afterward, the turning into the junctions lasts from frame 600 until frame 1700. This can be seen by the high amount of acceleration and yaw rate costs. When the turn is complete, the obstacle blocking the ego-vehicle lane is coming into focus. The required evasive maneuver causes deviation from the lane center and orientation. Due to this, the cost terms of orientation and driving area increase. When approaching the obstacle, the obstacle avoidance cost also rises. As the acceleration cost term shows, the ego-vehicle wants to increase speed when the obstacle is passed at frame 2800. Even in this complex scenario, our PSO planner can resolve this complex balancing act. Additionally, the fact that the cost term diagram allows the inference of the executed scenario demonstrates the appropriateness of the chosen cost terms. 

Looking into the valid particles after each optimization step is also insightful. This is done in the Fig.~\ref{fig:neckar_timing} and Fig.~\ref{fig:freiburger_timing}. As can be seen, in the straight-driving scenario, most of the particles are valid after the optimization step. This shows that our guided sampling strategy in the control space is effective. Even in a complex scenario, we can find valid particles and guarantee the safe operation of autonomous vehicles on public roads. This was also already demonstrated in many different real-world applications, such as the EVA-Shuttle project \cite{EVA-shuttle-hp} and the SHOW project \cite{show-hp}, where passenger transport with our autonomous shuttle busses was conducted. The primary assessment of the planning process takes place predominantly in and around Karlsruhe, Germany, encompassing both urban and suburban settings. Here, our planner was challenged with various significant challenges like navigating through narrow streets, coping with parked vehicles along the roads, negotiating intersections with limited visibility, and interacting with Vulnerable Road Users. During the testing and demonstration period, the shuttles drove more than 3.500\,km with the PSO planning algorithm. To be highlighted are the variants of the planner presented in this work were already used in several of our previous works (see, e.g.  \cite{schoerner2020optimization,schoerner2022risk, hubschneider2017a}) and proved to be reliable and easily extendable.

\newpage

\section{CONCLUSIONS}
\label{sec:conclusion}

In this paper, we presented a PSO-based motion planning approach that is utilized and applied in the field of automated driving. Due to our adaptations regarding computational efficiency, it is also able to be deployed on autonomous mobile platforms. The cost function is designed with both safety and comfort aspects in mind. The optimization strategy was further enhanced through advanced initialization and search methods. Additionally, we presented polar control space to yield accurate interpolations. The resulting planning algorithm was highly modular, flexible, and fast. Our approach was thereby able to handle complex, multi-modal, and dynamic scenarios. To prove that, the overall planning architecture was tested in real-world applications with more than 3.500\,km driven autonomously.

\section*{Acknowledgment}

This research was partially supported by Schwarz Mobility Solutions GmbH.

\printbibliography

\end{document}